\documentclass[sigconf, nonacm]{acmart}
\usepackage{float} 
\usepackage[toc]{appendix}

\AtBeginDocument{%
  \providecommand\BibTeX{{%
    \normalfont B\kern-0.5em{\scshape i\kern-0.25em b}\kern-0.8em\TeX}}}

\setcopyright{acmlicensed}
\copyrightyear{2024}
\acmYear{2024}
\acmDOI{XXXXXXX.XXXXXXX}

\acmConference[Conference acronym 'XX]{Make sure to enter the correct
  conference title from your rights confirmation email}{June 03--05,
  2018}{Woodstock, NY}
\acmISBN{978-1-4503-XXXX-X/18/06}

\graphicspath{{./images/}}

\begin{document}

\title{A Comparative Study of DSPy Teleprompter Algorithms for Aligning Large Language Models Evaluation Metrics to Human Evaluation}

\author{Bhaskarjit Sarmah}
\email{bhaskarjit.sarmah@blackrock.com}
\affiliation{%
  \institution{BlackRock}
  \city{Gurgaon}
  \country{India}
}

\author{Kriti Dutta}
\email{kriti.dutta@blackrock.com}
\affiliation{%
  \institution{BlackRock}
  \city{Gurgaon}
  \country{India}
}

\author{Anna Grigoryan}
\email{anna.grigoryan@blackrock.com}
\affiliation{%
  \institution{BlackRock}
  \city{Budapest}
  \country{Hungary}
}

\author{Sachin Tiwari}
\email{sachin.tiwari@blackrock.com}
\affiliation{%
  \institution{BlackRock}
  \city{Gurgaon}
  \country{India}
}

\author{Stefano Pasquali}
\email{stefano.pasquali@blackrock.com}
\affiliation{%
  \institution{BlackRock}
  \city{New York}
  \country{US}
}

\author{Dhagash Mehta}
\email{dhagash.mehta@blackrock.com}
\affiliation{%
  \institution{BlackRock}
  \city{New York}
  \country{US}
}

\renewcommand{\shortauthors}{Sarmah, et al.}

\begin{abstract}
We argue that the Declarative Self-improving Python (DSPy) optimizers are a way to align the large language model (LLM) prompts and their evaluations to the human annotations. We present a comparative analysis of five teleprompter algorithms, namely, Cooperative Prompt Optimization (COPRO), Multi-Stage Instruction Prompt Optimization (MIPRO), BootstrapFewShot, BootstrapFewShot with Optuna, and K-Nearest Neighbor Few Shot, within the DSPy framework with respect to their ability to align with human evaluations. As a concrete example, we focus on optimizing the prompt to align hallucination detection (using LLM as a judge) to human annotated ground truth labels for a publicly available benchmark dataset. Our experiments demonstrate that optimized prompts can outperform various benchmark methods to detect hallucination, and certain telemprompters outperform the others in at least these experiments.
\end{abstract}

\maketitle

\section{Introduction}
In the rapidly evolving field of generative artificial intelligence (GenAI), and more specifically large language models (LLMs), optimizing the input instructions to the LLMs \cite{naveed2023comprehensive}, also called 'prompts', has become one of the crucial areas of research and development \cite{sahoo2024systematic}. Optimizing prompts directly impacts the quality, accuracy, and efficiency of outputs generated by the LLMs; ensures that they align with task-specific requirements and data constraints; and, reduces reliance on manual trial-and-error approaches. 

We note that a similar problem is encountered even in traditional machine learning paradigm with tabular data, where certain datasets often come with expert-given labels, but their alignment with unsupervised clustering outcomes can vary depending on feature selection and distance metrics \cite{castellanos2024can}: if human-annotated ground-truth labels are available for the training dataset and an alignment with these labels is desired, then a supervised or semi-supervised learning approach is required rather than an unsupervised approach that does not use the ground-truth labels at all during the training.  

In Refs.~\cite{sudjianto2024human, sarmah2024choosethresholdevaluationmetric}, a post hoc approach to align human-annotated ground-truth labels to a corresponding LLM evaluation metric proposed where once the chosen LLM evaluation metric is computed on all the training samples, another linear or non-linear model is trained to learn the mapping between the LLM-computed scores and the human-annotated labels. In the present work, we propose to optimize the LLM as a judge prompt itself with respect to the ground-truth labels.

Declarative Self-improving Python (DSPy) \cite{khattab2023dspy} is a recently proposed framework that abstracts LLM pipelines into declarative modules which enable systematic optimization of these pipelines with respect to a chosen objective (e.g., accuracy of reproducing a given target variable). DSPy introduces a set of prompt optimizers, called \textit{teleprompters}, which systematically refine prompts with the objective being maximizing the performance of the LLM on the given specific tasks and datasets. In other words, these teleprompters automate the process of experimenting with prompt variations and evaluating their effectiveness against predefined metrics on the given training dataset and roughly mimic the supervised learning process as in the traditional machine learning paradigm.

We emphasis that DSPy and model fine-tuning \cite{han2024parameter,ling2023domain} are alternative approaches to optimizing LLMs for a given task. DSPy focuses on optimizing prompts and pipeline configurations through declarative modules and automated teleprompters without modifying the pre-trained LLM's weights. Because the weight parameters of the model are not tuned, this approach is relatively faster and cost-effective as it requires less computational resources. On the other hand, fine-tuning involves updating the LLM's weights using task-specific training data and may yield more robust performance for narrowly defined applications but is usually computationally more intensive. It also requires significant amount of labeled data, and may sometimes lead to over-fitting. In the present work, we focus only on teleprompters for prompt optimization, and the fine-tuning and combination between the two will be discussed in a future work.

A key innovation of teleprompters lies in their ability to leverage few-shot prompting \cite{sahoo2024systematic} which is a prompting technique where a prompt is supplemented with a few example pairs of inputs and desired outputs to guide the LLM's response. Few-shot prompting has been shown to substantially improve LLM performance across diverse tasks by enabling models to better align their outputs with task-specific requirements. Teleprompters automate the selection and incorporation of high-quality examples, further refining the optimization process to produce more accurate and contextually aligned outputs.

Various teleprompter algorithms have been proposed in the literature, and by systematically analyzing the strengths and weaknesses of different algorithms, we can uncover valuable insights into their optimization capabilities, leading to more effective and efficient DSPy pipeline for specific tasks. The primary objective of this research paper is to conduct a comprehensive evaluation of five DSPy teleprompter algorithms. We compare their performances and assess their ability to enhance alignment between the context-answer pairs (and also question-context-answer triplets) for the given questions and the ground-truth annotations, with the help of a recently published public benchmark dataset for hallucination detection, called HaluBench \cite{ravi2024lynx}. In turn, for each teleprompter, we identify corresponding optimized prompts to compute the faithfulness score for the retrieval augmented generation (RAG) system underneath the benchmark dataset.

Note that Ref.\cite{wan2024teach} comes closer to our work where they compare and contrast two types of strategies for prompt optimizations and their combinations: examplar optimization and instruction optimization. However, the present work focuses only on a comparative study for teleprompter optimization, and is a complementary study to the former. Similarly, Ref.~\cite{yudspy} proposes to use DSPy to optimize prompts to detect jailbreak prompts, which in turn optimize guardrails for jailbreak through self-improving language model pipelines. However, this paper does not provide any comparative or extensive study on teleprompters.

The structure of this paper is as follows: Section \ref{sec:dspy} introduces the DSPy framework and the concept of teleprompters. Section \ref{sec:datasets} provides details on the public benchmark dataset and preprocessing of the data used in the present study. Section \ref{sec:methodology} outlines the methodology, including the evaluation metrics used in the study and computational details. Section \ref{sec:results} discusses the implications of the findings and highlights potential areas for improvement. Finally, Section \ref{sec:conclusion} concludes the paper with a summary of key insights and directions for future research.

\section{DSPy Teleprompters}\label{sec:dspy}
DSPy\cite{khattab2023dspy} is a programming model that is designed to optimize the prompts in a more structured (specifically, programmatic) and efficient manner. To set up some terminologies, let's first define \textit{modules} as small units or building blocks of an abstracted LLM pipeline and are reusable components that may be used for different tasks. A \textit{signatures} is a short declarative specification that informs what a text transformation module should take as an input and produce as an output (e.g., the signature "question -> answer" means that the input will be a question in the text format and output will be an answer in the text format).

Then, a module called \textit{predictor} DSPy module that generates an output from an input based on a given signature that uses an LLM to "predict" the answer (e.g., for a given question, it will predict an answer using an LLM provided the signature is "question -> answer"). Similarly, the \textit{adapter} module transforms the input to a desired format that is suitable for a different kind of module or a step in the pipeline. This module is used for tasks like formatting, changing data types, adding additional information, etc. One can also device modules such as \textit{ChainOfThought} that takes a question, thinks step-by-step, and gives an answer using an LLM. One can also such a module with another module, for example, ChainOfThought can explain the reasoning before giving an answer from another module.

An \textit{assertions} is a condition (or, a set of conditions) that must be true during pipeline execution (e.g., ensuring that an output is not empty). A \textit{metric} is an objective evaluation metric (e.g., exact match between two sentences) to quantitatively measure the performance of the LLM for the given data for the given task. To optimize a given module, modules or the entire pipeline, DSPy uses an \textit{optimizer} that tunes parameters such as which examples are used for training or which prompts work best.

A special type of optimizer in DSPy is called a \textit{teleprompter} which is an optimizer that further improves the quality of modules via prompting or fine-tuning. It follows the below steps to optimize the modules: 
\begin{enumerate}
\item \textbf{Candidate Generation:} As the first step, the teleprompter starts by finding all instances of the Predictor modules in a given DSPy program and generates potential candidates for parameters (e.g., prompt templates, instructions, field descriptions, demonstrations for input-output examples, etc.) using simulation or sampling from a teacher model to create new examples. E.g., for a module with a signature "question -> answer", the teleprompter will run the module multiple times with different prompts and gather the output as potential examples.

\item \textbf{Parameter Optimization:} Now, the teleprompter optimizes the candidate parameters using methods such as random search or hyperparameter tuning, and selects the best combination of parameters by running trials and evaluating based on a specific metric. This stage might also involve fine-tuning a smaller language model based on the gathered examples to ensure a better fit. E.g., the teleprompter might evaluate different prompt examples with an accuracy metric to see which set provides the highest accuracy for answering questions. If fine-tuning is being used, it might update the language model weights with the best examples to improve consistency.

\item \textbf{Higher-Order Program Optimization:} Once the optimized module parameters are obtained, the teleprompter may further optimize the pipeline by changing its overall structure by, for example, combining multiple copies of the optimized program into an ensemble. This stage may also introduce advanced strategies such as automatic backtracking or conditional logic to improve the robustness and adaptability of the pipeline. E.g., if multiple versions of a question-answering pipeline exist, the teleprompter could create an ensemble of these versions that then aggregates their outputs to get a more reliable answer.
\end{enumerate}

In the present study, we employ a systematic approach to analyze the performance of the five DSPy teleprompter algorithms that are available in the literature along with publicly available implementations: COPRO, MIPRO, BootstrapFewShot, BootstrapFewShot with Optuna, and KNN Few Shot. Below is a brief description of each algorithm:

\noindent\textbf{Cooperative Prompt Optimization (COPRO)}: The COPRO algorithm is designed to optimize prompts by leveraging a cooperative approach that systematically explores various prompt configurations and evaluates their performance based on predefined metrics. The algorithm iteratively refines the prompts to achieve the best possible outcomes, making it a robust tool for enhancing the accuracy and efficiency of DSPy programs.

\noindent\textbf{Multi-Stage Instruction Prompt Optimization (MIPRO)}: MIPRO \cite{opsahl2024optimizing} is a novel optimizer that focuses on improving instructions for LLMs by optimizing the instructions in multiple stages. MIPRO is particularly effective in scenarios where complex instructions are required, making it a valuable addition to the DSPy framework.

\noindent\textbf{Bootstrap Few-Shot}: The Bootstrap Few-Shot \cite{khattab2023dspy} algorithm extends the signature by automatically generating and including optimized examples within the prompt sent to the model. This approach, known as few-shot learning, involves providing example model inputs and desired outputs within the prompt. By incorporating these examples, Bootstrap Few-Shot enhances the model's performance on a wide range of tasks.

\noindent\textbf{Bootstrap Few-Shot with Optuna}: This \cite{khattab2023dspy} variant of the Bootstrap Few-Shot algorithm integrates Optuna \cite{akiba2019optuna}, a hyperparameter optimization framework. Optuna employs state-of-the-art algorithms for sampling hyperparameters and efficiently pruning unpromising trials. By combining the few-shot learning capabilities of Bootstrap Few-Shot with the hyperparameter optimization techniques of Optuna, this algorithm achieves superior performance and efficiency.

\noindent\textbf{K-nearest Neighbor Few Shot}: The KNN Few Shot \cite{khattab2023dspy} algorithm utilizes the KNN approach to select few-shot examples. By identifying and incorporating the most relevant examples, this algorithm may significantly boost the performance of DSPy programs. The KNN Few Shot algorithm is particularly effective in scenarios where the selection of high-quality examples is crucial for optimizing the model's performance.

These five algorithms represent a diverse set of optimization strategies, each with its unique strengths and applications. The subsequent sections of this paper will provide a detailed analysis of their performance scores before and after compilation, offering valuable insights into their optimization capabilities.

\section{Datasets}\label{sec:datasets}
We use the recently publicly released HaluBench\footnote{\url{https://huggingface.co/datasets/PatronusAI/HaluBench}} dataset to evaluate the performance of all DSPy teleprompter algorithms. HaluBench is an open-source hallucination evaluation benchmark dataset comprising around 15K Context-Question-Answer triplets as well as human-annotated ground-truth labels 'Pass' (i.e., not hallucinated) and 'Fail' (hallucinated). It is the first benchmark to include hallucination tasks from real-world domains like finance, medicine, etc. The dataset was constructed by sourcing examples from publicly available datasets, including FinanceBench, PubmedQA, CovidQA, HaluEval, DROP, and RAGTruth, which we call sub-datasets. By providing a diverse set of real-world scenarios, HaluBench enables comprehensive evaluation of hallucination detection in language models.

For our purposes, the presence of context and ground truth answers is crucial as it allows for a more accurate assessment of their optimization capabilities. The RAG system (which is not available in the dataset though) used to generate the answers in this dataset may leverage the provided context to generate answers, enabling a comprehensive evaluation of the algorithms' performance in a controlled environment.

\section{Methodology and Computational Details}\label{sec:methodology}
In this Section, we describe the methodology of our experiments and also provide computational details for reproducibility purposes.

\subsection{Dataset Preparation}

To ensure high-quality data for evaluating faithfulness, we applied several data-cleaning and sampling steps across six sub-datasets: \texttt{CovidQA}, \texttt{FinanceBench}, \texttt{HaluEval}, \texttt{PubMedQA}, \texttt{DROP}, and \texttt{RAGTruth}.

\subsubsection{Data Cleaning}

As part of data cleaning, we removed any sample where the ground-truth answer contained three or fewer tokens, as these samples were deemed insufficient for meaningful evaluation (e.g., extracting 'statements' or 'claims', that is required by many methodologies faithfulness scores, from such short answers is not feasible nor meaningful in most cases.). This filtering step reduced the dataset to a total of 9,616 samples, ensuring that the retained samples contained substantive responses for a more accurate assessment of model faithfulness.

Due to computational limitations, we further reduced the dataset by applying \textit{stratified sampling}, ensuring proportional representation across both the six sub-datasets and the binary classes. Eventually, we focused on a total of 1,500 samples, distributed into three subsets: training set with 750 samples, validation Set with 375 samples and test set with 375 samples. Stratification also ensured that each sub-dataset and both target classes were proportionally represented in all three subsets. This approach preserved the original data distribution while maintaining fairness across datasets and binary classes, supporting robust and consistent evaluation across all stages of our experiments.

Inspired by the prompt format proposed in Ref.~\cite{ravi2024lynx}, we adapted the input structure to exclude the \textit{QUESTION} component, focusing solely on the \textit{DOCUMENT} (i.e., context) and \textit{ANSWER}. This was done to evaluate faithfulness independently of question semantics.

\subsection{DSPy Optimizers and Configurations}
We utilized various DSPy optimizers\footnote{\url{https://github.com/stanfordnlp/dspy}}, each with tailored hyperparameters for faithfulness evaluation tasks as summarized in Table \ref{tab:dspy-configurations}.

\begin{table}[ht]
\centering
\small 
\caption{DSPy Optimizers and Their Hyperparameter Configurations}
\label{tab:dspy-configurations}
\begin{tabular}{p{2.8cm}p{5.2cm}}
\toprule
\textbf{DSPy Optimizer} & \textbf{Hyperparameters} \\
\midrule
Baseline CoT & None \\
\midrule
Bootstrap Few Shot (8 Examples) & \texttt{max\_bootstrapped\_demos=8}, \texttt{max\_labeled\_demos=16} \\
\midrule
Bootstrap Few Shot Random Search & \texttt{max\_bootstrapped\_demos=8}, 
\texttt{max\_labeled\_demos=16},
\texttt{num\_candidate\_programs=3} \\
\midrule
Bootstrap Few Shot Optuna & \texttt{max\_bootstrapped\_demos=8}, 
\texttt{max\_labeled\_demos=16},
\texttt{num\_candidate\_programs=3} \\
\midrule
MIPROv2 & \texttt{num\_candidates=3}, \texttt{init\_temperature=0}, \texttt{max\_bootstrapped\_demos=8}, \texttt{max\_labeled\_demos=16},
\texttt{num\_trials\=15}, \texttt{minibatch\_size=25}, \texttt{minibatch\_full\_eval\_steps=10}
\\
\midrule
KNN Few Shot & \texttt{K=8} \\
\midrule
COPRO & \texttt{max\_bootstrapped\_demos=8}, 
\texttt{max\_labeled\_demos=16}, \texttt{breadth=5}, \texttt{depth=3}, \texttt{init\_temperature=0} \\
\bottomrule
\end{tabular}
\end{table}

\subsection{Experimental Workflow}
\label{sec:expworkflow}

We conducted experiments using OpenAI’s \texttt{GPT-4o} model as the baseline LLM. To ensure experimental consistency and reproducibility, the following hyperparameters were fixed across all evaluations:

\begin{table}[H] 
\centering
\caption{Hyperparameter Configuration for GPT-4o}
\label{tab:hyperparameters}
\resizebox{\columnwidth}{!}{%
\begin{tabular}{ll}
\toprule
\textbf{Hyperparameter} & \textbf{Value / Description} \\
\midrule
Temperature     & 0.0 \\
Maximum Tokens  & None\\
Top-p Sampling  & 1.0\\
Output Format   & JSON format with \texttt{REASONING} and \texttt{SCORE} keys \\
\bottomrule
\end{tabular}%
}
\end{table}

\subsubsection{Baseline Prompt for GPT-4o}
To establish the baseline performance of \texttt{GPT-4o}, we used the following prompt:

\small
\begin{verbatim}
Given the following DOCUMENT and ANSWER, 
determine whether the ANSWER is faithful to 
the contents of the DOCUMENT. The ANSWER 
must not offer new information beyond the 
context provided in the DOCUMENT. It must 
also not contradict the DOCUMENT.

DOCUMENT:
{context}

ANSWER:
{answer}

Your output should be in JSON format with 
the keys:
"REASONING" and "SCORE".

{
  "REASONING": <your reasoning as bullet points>,
  "SCORE": <"PASS" if the answer is faithful, 
            "FAIL" if it is not>
}
\end{verbatim}

\subsubsection{Experimental Pipeline}
The evaluation pipeline consisted of four sequential phases, integrating both baseline and optimized evaluations:

\begin{enumerate}
    \item \textbf{Baseline Evaluation:} We first evaluated GPT-4o on the test set using the aforementioned baseline prompt without applying any DSPy optimizers or teleprompters. This established benchmarks for key metrics such as exact match accuracy and F1 scores, serving as the reference for future comparisons.

    \item \textbf{DSPy Optimization:} DSPy optimizers were applied on the training set to refine model prompts and adjust relevant hyperparameters. This optimization process aimed to enhance both prediction accuracy and class-level balance.

    \item \textbf{Validation Phase:} The optimized prompts were applied to the validation set to tune and adjust optimizer-specific hyperparameters (e.g., the number of few-shot examples, learning rates); and, to ensure robustness by evaluating generalization performance on unseen validation data, mitigating overfitting risks during training.

    \item \textbf{Testing Phase:} After hyperparameter tuning, we evaluated the final optimized prompts on the test set using exact match accuracy, micro F1, macro F1, and weighted-macro F1 scores. All evaluations were conducted across datasets, including \texttt{CovidQA}, \texttt{FinanceBench}, \texttt{HaluEval}, \texttt{PubMedQA}, \texttt{DROP}, and \texttt{RAGTruth}.

    \item \textbf{Public Faithfulness Benchmark Comparison:} For further comparison with publicly available implementations of faithfulness evaluation, we used corresponding functions from \texttt{RAGAS}\footnote{\url{https://github.com/explodinggradients/ragas}} \cite{es2023ragas} and \texttt{DeepEval}\footnote{\url{https://docs.confident-ai.com/docs/metrics-hallucination}}. Their default implementations, with a threshold value of 0.5 for faithfulness scores\footnote{Note that, arguably, using default values of the implementations for specific dataset with ground truth label. However, identifying an appropriate threshold for the given data is beyond the scope of this work, and the reader is referred to our companion work. In the present work, the default thresholds are used as a general purpose benchmark only.}, were applied exclusively to the test dataset.
\end{enumerate}

\subsubsection{Evaluation Metrics}
Since the datasets are highly imbalanced across different sub-datasets, for evaluating different prompts, we used the micro, macro and weighted macro F1 metrics. To gain deeper insights into model performance, we further analyzed the F1 scores at a sub-group level for each sub-dataset in the test set. This breakdown helped identify domain-specific trends, such as variations in performance across datasets like FinanceBench, PubmedQA, or DROP.

\section{Results}
\label{sec:results}

The effectiveness of DSPy optimizers was evaluated in improving hallucination detection (as annotated in the ground-truth labels in the HaluBench dataset) using OpenAI’s GPT-4o model as a base evaluator. We report two sets of results: aggregated metrics across all datasets, highlighting overall performance; and, detailed dataset evaluations are presented in Appendix~\ref{sec:appendix}, where we explore individual dataset performance and specific trends. The primary evaluation metrics are Micro, Macro, and Weighted F1 scores, which measure the ability of models to balance overall accuracy, minority class detection, and proportional representation. Additionally, accuracy is reported as it was used as the objective for training the optimizers.

Tables~\ref{tab:accuracy_scores} and~\ref{tab:f1_scores} present the aggregate level results for the hold-out test set.

\begin{table}[ht]
\centering
\caption{Overall Accuracy Scores on the entire test set (DSPy Metric)}
\label{tab:accuracy_scores}
\resizebox{\columnwidth}{!}{%
\begin{tabular}{lc}
\toprule
\textbf{Model/Optimizer} & \textbf{Accuracy (\%)} \\
\midrule
Baseline GPT-4o                    & 80.91 \\
Baseline CoT DSPy                  & 82.13 \\
Bootstrap Few Shot (8 Examples)    & 81.60 \\
Bootstrap Few Shot Random Search   & 84.00 \\
Bootstrap Few Shot Optuna          & 85.60 \\
MIPROv2                            & \textbf{85.87} \\
KNN Few Shot                       & 83.47 \\
COPRO                             & 82.13 \\
RAGAS                              & 61.60 \\
DeepEval                           & 61.60 \\
\bottomrule
\end{tabular}%
}
\end{table}

\begin{table}[ht]
\centering
\caption{Total F1 Scores (Micro, Macro, Weighted)}
\label{tab:f1_scores}
\resizebox{\columnwidth}{!}{%
\begin{tabular}{lccc}
\toprule
\textbf{Model/Optimizer} & \textbf{Micro F1} & \textbf{Macro F1} & \textbf{Weighted F1} \\
\midrule
Baseline GPT-4o                    & 0.8091  & 0.8019  & 0.8125  \\
Baseline CoT DSPy                  & 0.792   & 0.2267  & 0.8008  \\
Bootstrap Few Shot (8 Examples)    & 0.760   & 0.3886  & 0.7868  \\
Bootstrap Few Shot Random Search   & 0.816   & \textbf{0.8115}  & 0.8197  \\
Bootstrap Few Shot Optuna          & 0.8053  & 0.8006  & 0.8092  \\
MIPROv2                            & \textbf{0.8187}  & 0.4082  & \textbf{0.8248}  \\
KNN Few Shot                       & 0.7754  & 0.3883  & 0.7844  \\
COPRO                              & 0.792   & 0.2267  & 0.8008  \\
RAGAS                              & 0.616   & 0.5663  & 0.6074  \\
DeepEval                           & 0.667   & 0.625   & 0.625   \\
\bottomrule
\end{tabular}%
}
\end{table}

\subsection{Key Insights and Trends}

Baseline GPT-4o demonstrated strong overall performance but faced challenges in maintaining class-level balance, as evidenced by its low Macro F1 scores. In contrast, the best-performing approaches showcased remarkable improvements: MIPROv2 achieved the highest weighted F1 score of 0.8248, leveraging structured optimization techniques to enhance overall accuracy. Meanwhile, Few Shot Random Search stood out with the highest Macro F1 score of 0.8115, reflecting its ability to effectively handle minority classes and ensure more balanced performance across categories.

Performance varied notably across different datasets, highlighting the influence of dataset characteristics on optimization outcomes. Structured datasets such as \texttt{CovidQA} and \texttt{PubMedQA} appeared to benefit most from optimization strategies, achieving robust results due to their consistent patterns. However, datasets with greater variability, such as \texttt{FinanceBench}, can present unique challenges that may require more adaptive prompt engineering techniques to address their diverse and less predictable structures effectively.

Detailed evaluations for each dataset are reported in Appendix: optimization approaches, such as MIPROv2 and Few Shot Random Search, excelled in structured datasets like \texttt{CovidQA} and \texttt{PubMedQA}, achieving high macro F1 scores. These results highlight their ability to pick prompts that effectively detect hallucinations within well-defined contexts, where structured patterns and clear task objectives enhance model performance. Such datasets benefit significantly from optimization techniques, showcasing the importance of alignment between dataset characteristics and model strategies.

In contrast, unstructured datasets like \texttt{FinanceBench} and \texttt{DROP} posed greater challenges for hallucination detection. Baseline models struggled due to the variability and complexity inherent in these datasets, while optimization methods only partially mitigated these issues, indicating the need for further refinement. Among these methods, Few Shot Random Search consistently demonstrated high Macro F1 scores across both structured and unstructured datasets, underscoring its effectiveness in improving sensitivity to minority classes, even in diverse and less predictable environments.

\section{Results Discussion}
\label{sec:discussion}

The results demonstrate that while baseline GPT-4o achieves reasonable performance, it struggles with class imbalance, as evidenced by its low macro F1 scores. This underscores the need for tailored optimization frameworks to improve sensitivity to minority classes.

\paragraph{Class-Level Balance:}
Optimizers such as Bootstrap Few Shot with Random Search and with Optuna improved minority class detection by balancing class-specific F1 scores. Bootstrap Few Shot with Random Search was particularly effective due to its exploration of diverse prompt configurations.

\paragraph{Batch and Structured Learning:}
MIPROv2 achieved the best Weighted F1 score by leveraging structured learning techniques and batching. This method balanced both class-specific sensitivity and global accuracy.

\paragraph{Prompt-Specific Adaptations:}
COPRO and KNN Few Shot struggled with datasets like \texttt{FinanceBench}, where prompt-specific adaptations were less effective due to the dataset's variable structure.

\paragraph{Structured Datasets:}
Optimizers excelled in datasets with well-defined contexts, such as \texttt{CovidQA} and \texttt{PubMedQA}, achieving high Macro F1 scores due to clear prompt-document-answer alignment.

\paragraph{Unstructured Datasets:}
Datasets like \texttt{DROP} and \texttt{FinanceBench} presented more complex inputs, where hallucination detection proved more challenging due to fragmented context and high answer variability.

\paragraph{Teleprompter Optimization Analysis:}
During teleprompter optimization, the DSPy routines select examples used to guide the model's output based on maximizing the overall evaluation metric, specifically exact match across the entire dataset. However, the specific dataset at hand in the present work consists of multiple heterogeneous sub-datasets arising from diverse sources. While optimizing on the overall dataset produced strong aggregate results, a detailed analysis of data source-level performance revealed inconsistencies. In particular, some data sources exhibited significantly lower performance, despite high aggregate evaluation scores. i.e., the teleprompters optimized for maximum overall performance may rely heavily on examples from only a subset of the data sources, specifically those where the model already performed well. 

Since the optimization process focuses on selecting the most impactful examples to increase the evaluation score, this process may inadvertently overfit to high-performing data sources while neglecting examples from lower-performing sources. As a result, the optimized prompts may fail to generalize effectively across all data sources such as \texttt{FinanceBench} sub-dataset in our case.

To address this limitation, we propose using a stratified sampling approach \textit{during} teleprompter optimization in the future. This approach would ensure that examples selected for prompt optimization are proportionally sampled from each data source, rather than solely focusing on maximizing the overall evaluation metric. By ensuring that each data source contributes representative examples, the teleprompters could be made more robust, improving generalization and reducing bias toward specific data types. Instead of selecting only the "best" examples that maximize the overall performance, the optimization process should prioritize diversity by ensuring that examples from underrepresented or challenging data sources are also included.

\paragraph{Potential Model Bias due to Public Data Exposure:}
An additional consideration relates to the model's observed performance on publicly available datasets. While we cannot definitively determine GPT-4's pretraining data, its strong results on certain datasets such as \texttt{PubMedQA} and \texttt{CovidQA} suggest potential prior exposure. Since these datasets are publicly accessible, it is plausible that they could have influenced the pretraining process, either directly or indirectly.

This possibility highlights the need for careful evaluation and dataset selection in future research. If public datasets inadvertently align with the model's training corpus, they may yield inflated performance metrics, masking potential weaknesses when handling unseen or proprietary data sources. Evaluating models on more controlled and proprietary datasets could mitigate this risk and provide a clearer picture of generalization performance in real-world applications.

\section{Conclusion}\label{sec:conclusion}
This study systematically evaluated DSPy optimizers for improving hallucination detection using GPT-4o as a base evaluator. Our experiments yield that baseline GPT-4o delivers strong overall performance but exhibits limitations in sensitivity to minority classes, as evidenced by its low Macro F1 scores. This shortfall highlights its difficulty in achieving balanced performance across all categories, particularly when addressing nuanced or less frequent patterns. These challenges underscore the need for targeted optimization to improve class-level balance.

Optimizers such as Bootstrap Few Shot Random Search and MIPROv2 effectively address these limitations, demonstrating significant advancements in hallucination detection. Their performance is particularly pronounced in structured datasets, where clear patterns enable optimization techniques to achieve robust results. However, the variability and complexity of unstructured datasets present ongoing challenges, emphasizing the critical role of dataset structure in determining the success of optimization strategies.

The study highlights the need for adaptive optimizers, dynamic prompt generation, and dataset-specific evaluation frameworks and demonstrates the importance of combining optimization with advanced techniques like instruction fine-tuning to enhance hallucination detection across diverse datasets.

Building on these findings, future research should prioritize the combined use of instruction fine-tuning and prompt optimization to enhance language models' adaptability across diverse datasets. This integrated approach aligns model outputs more closely with specific tasks, leading to improved performance in both structured and unstructured contexts~\cite{soylu2024fine}. Moreover, an optimized prompting strategy for hallucination detection and other metrics may provide cleaner distributions of the corresponding scores with respect to the ground-truth labels and, in turn, crisper thresholds for the corresponding evaluation metrics \cite{sarmah2024choosethresholdevaluationmetric}.

Beyond hallucination detection, the DSPy techniques can also scale to address other critical evaluation tasks, such as reasoning, toxicity, and bias assessment. Extending these methods will contribute to the broader goal of enhancing LLM reliability and robustness, ensuring their effectiveness across a wider range of applications. By addressing these areas, future advancements can significantly improve both the adaptability and trustworthiness of language models in diverse operational settings.


\section*{Acknowledgement}
The views expressed here are those of the authors alone and not of BlackRock, Inc.

\bibliographystyle{unsrt}
\bibliography{references.bib}

\begin{thebibliography}{10}

\bibitem{naveed2023comprehensive}
Humza Naveed, Asad~Ullah Khan, Shi Qiu, Muhammad Saqib, Saeed Anwar, Muhammad Usman, Naveed Akhtar, Nick Barnes, and Ajmal Mian.
\newblock A comprehensive overview of large language models.
\newblock {\em arXiv preprint arXiv:2307.06435}, 2023.

\bibitem{sahoo2024systematic}
Pranab Sahoo, Ayush~Kumar Singh, Sriparna Saha, Vinija Jain, Samrat Mondal, and Aman Chadha.
\newblock A systematic survey of prompt engineering in large language models: Techniques and applications.
\newblock {\em arXiv preprint arXiv:2402.07927}, 2024.

\bibitem{castellanos2024can}
Nathalia~M Castellanos, Sebastian Frank, Dhruv Desai, Stefano Pasquali, and Dhagash Mehta.
\newblock Can an unsupervised clustering algorithm reproduce a categorization system?
\newblock In {\em Proceedings of the 5th ACM International Conference on AI in Finance}, pages 213--221, 2024.

\bibitem{sudjianto2024human}
Agus Sudjianto and Srinivas Neppalli.
\newblock Human-calibrated automated testing and validation of generative language models: An overview.
\newblock {\em Available at SSRN}, 2024.

\bibitem{sarmah2024choosethresholdevaluationmetric}
Bhaskarjit Sarmah, Mingshu Li, Jingrao Lyu, Sebastian Frank, Nathalia Castellanos, Stefano Pasquali, and Dhagash Mehta.
\newblock How to choose a threshold for an evaluation metric for large language models, 2024.

\bibitem{khattab2023dspy}
Omar Khattab, Arnav Singhvi, Paridhi Maheshwari, Zhiyuan Zhang, Keshav Santhanam, Sri Vardhamanan, Saiful Haq, Ashutosh Sharma, Thomas~T Joshi, Hanna Moazam, et~al.
\newblock Dspy: Compiling declarative language model calls into self-improving pipelines.
\newblock {\em arXiv preprint arXiv:2310.03714}, 2023.

\bibitem{han2024parameter}
Zeyu Han, Chao Gao, Jinyang Liu, Jeff Zhang, and Sai~Qian Zhang.
\newblock Parameter-efficient fine-tuning for large models: A comprehensive survey.
\newblock {\em arXiv preprint arXiv:2403.14608}, 2024.

\bibitem{ling2023domain}
Chen Ling, Xujiang Zhao, Jiaying Lu, Chengyuan Deng, Can Zheng, Junxiang Wang, Tanmoy Chowdhury, Yun Li, Hejie Cui, Xuchao Zhang, et~al.
\newblock Domain specialization as the key to make large language models disruptive: A comprehensive survey.
\newblock {\em arXiv preprint arXiv:2305.18703}, 2023.

\bibitem{ravi2024lynx}
Selvan~Sunitha Ravi, Bartosz Mielczarek, Anand Kannappan, Douwe Kiela, and Rebecca Qian.
\newblock Lynx: An open source hallucination evaluation model.
\newblock {\em arXiv preprint arXiv:2407.08488}, 2024.

\bibitem{wan2024teach}
Xingchen Wan, Ruoxi Sun, Hootan Nakhost, and Sercan~O Arik.
\newblock Teach better or show smarter? on instructions and exemplars in automatic prompt optimization.
\newblock {\em arXiv preprint arXiv:2406.15708}, 2024.

\bibitem{yudspy}
Boxi Yu and Pinjia He.
\newblock Dspy guardrails: Building safe llm applications via self-refining language model pipelines.

\bibitem{opsahl2024optimizing}
Krista Opsahl-Ong, Michael~J Ryan, Josh Purtell, David Broman, Christopher Potts, Matei Zaharia, and Omar Khattab.
\newblock Optimizing instructions and demonstrations for multi-stage language model programs.
\newblock {\em arXiv preprint arXiv:2406.11695}, 2024.

\bibitem{akiba2019optuna}
Takuya Akiba, Shotaro Sano, Toshihiko Yanase, Takeru Ohta, and Masanori Koyama.
\newblock Optuna: A next-generation hyperparameter optimization framework.
\newblock In {\em Proceedings of the 25th ACM SIGKDD international conference on knowledge discovery \& data mining}, pages 2623--2631, 2019.

\bibitem{es2023ragas}
Shahul Es, Jithin James, Luis Espinosa-Anke, and Steven Schockaert.
\newblock Ragas: Automated evaluation of retrieval augmented generation.
\newblock {\em arXiv preprint arXiv:2309.15217}, 2023.

\bibitem{soylu2024fine}
Dilara Soylu, Christopher Potts, and Omar Khattab.
\newblock Fine-tuning and prompt optimization: Two great steps that work better together.
\newblock In {\em Proceedings of the 2024 Conference on Empirical Methods in Natural Language Processing}, pages 10696--10710. Association for Computational Linguistics, 2024.

\end{thebibliography}


\clearpage

\appendix
\setcounter{table}{0}
\renewcommand{\thetable}{A\arabic{table}}

\section{Results Breakdown by Sub-datasets}\label{sec:appendix}

This appendix presents detailed results for individual datasets, providing metrics such as Micro F1, Macro F1, and Weighted Macro F1 for PubMedQA, HaluEval, RAGTruth, DROP, CovidQA and FinanceBench sub-datasets in Tables \ref{tab:pubmedqa_results}, \ref{tab:halueval_results}, \ref{tab:ragtruth_results}, \ref{tab:drop_results}, \ref{tab:covidqa_results} and \ref{tab:financebench_results}, respectively. These dataset-level evaluations complement the overall results reported in the main paper.

\begin{table}[h]
\centering
\caption{\texttt{PubMedQA} Dataset Results}
\label{tab:pubmedqa_results}
\resizebox{\columnwidth}{!}{%
\begin{tabular}{lccc}
\toprule
\textbf{Model/Optimizer} & \textbf{Micro F1} & \textbf{Macro F1} & \textbf{Weighted F1} \\
\midrule
Baseline GPT-4o                    & 0.8205  & 0.8126  & 0.8136  \\
Baseline CoT DSPy                  & 0.8462  & 0.8452  & 0.8455  \\
Bootstrap Few Shot Random Search   & \textbf{0.9231}  & \textbf{0.9231}  & 0.9231  \\
Bootstrap Few Shot Optuna          & 0.8718  & 0.8704  & 0.8708  \\
MIPROv2                            & \textbf{0.9231}  & \textbf{0.9231}  & \textbf{0.9349}  \\
KNN Few Shot                       & 0.8462  & 0.4391  & 0.8771  \\
COPRO                              & 0.8462  & 0.8452  & 0.8455  \\
RAGAS                              & 0.8718  & 0.8704  & 0.8708  \\
DeepEval                           & 0.8205  & 0.5564  & 0.5755  \\
\bottomrule
\end{tabular}%
}
\end{table}

\begin{table}[h]
\centering
\caption{\texttt{HaluEval} Dataset Results}
\label{tab:halueval_results}
\resizebox{\columnwidth}{!}{%
\begin{tabular}{lccc}
\toprule
\textbf{Model/Optimizer} & \textbf{Micro F1} & \textbf{Macro F1} & \textbf{Weighted F1} \\
\midrule
Baseline GPT-4o                    & 0.8032  & 0.7766  & 0.8165  \\
Baseline CoT DSPy                  & 0.7976  & 0.7756  & 0.8119  \\
Bootstrap Few Shot Random Search   & \textbf{0.8095}  & \textbf{0.7857}  & \textbf{0.8222}  \\
Bootstrap Few Shot Optuna          & 0.7619  & 0.7305  & 0.7770  \\
MIPROv2                            & 0.8016  & 0.7781  & 0.8153  \\
KNN Few Shot                       & 0.7610  & 0.7332  & 0.7774  \\
COPRO                              & 0.7976  & 0.7756  & 0.8119  \\
RAGAS                              & 0.5635  & 0.3765  & 0.5527  \\
DeepEval                           & 0.5595  & 0.5564  & 0.5755  \\
\bottomrule
\end{tabular}%
}
\end{table}

\begin{table}[h]
\centering
\caption{\texttt{RAGTruth} Dataset Results}
\label{tab:ragtruth_results}
\resizebox{\columnwidth}{!}{%
\begin{tabular}{lccc}
\toprule
\textbf{Model/Optimizer} & \textbf{Micro F1} & \textbf{Macro F1} & \textbf{Weighted F1} \\
\midrule
Baseline GPT-4o                    & \textbf{0.9143}  & \textbf{0.8583}  & \textbf{0.9168}  \\
Baseline CoT DSPy                  & 0.8000  & 0.6694  & 0.8059  \\
Bootstrap Few Shot Random Search   & 0.8571  & 0.7304  & 0.8519  \\
Bootstrap Few Shot Optuna          & 0.8226  & 0.6983  & 0.8286  \\
MIPROv2                            & 0.8228  & 0.4807  & 0.8364  \\
KNN Few Shot                       & 0.8000  & 0.3741  & 0.7838  \\
COPRO                              & 0.8000  & 0.6694  & 0.8059  \\
RAGAS                              & 0.7714  & 0.4355  & 0.7217  \\
DeepEval                           & 0.8000  & 0.4440  & 0.7365  \\
\bottomrule
\end{tabular}%
}
\end{table}

\begin{table}[h]
\centering
\caption{\texttt{DROP} Dataset Results}
\label{tab:drop_results}
\resizebox{\columnwidth}{!}{%
\begin{tabular}{lccc}
\toprule
\textbf{Model/Optimizer} & \textbf{Micro F1} & \textbf{Macro F1} & \textbf{Weighted F1} \\
\midrule
Baseline GPT-4o                    & 0.2500  & 0.2448  & 0.2343  \\
Baseline CoT DSPy                  & 0.5000  & 0.4857  & 0.5000  \\
Bootstrap Few Shot Random Search   & 0.3333  & 0.3143  & 0.2952  \\
Bootstrap Few Shot Optuna          & 0.5833  & 0.5556  & 0.5741  \\
MIPROv2                            & 0.5833  & 0.5556  & 0.5741  \\
KNN Few Shot                       & 0.4167  & 0.4126  & 0.5000  \\
COPRO                              & 0.5000  & 0.4857  & 0.5000  \\
RAGAS                              & 0.4167  & 0.2941  & 0.2451  \\
DeepEval                           & 0.5833  & 0.3684  & 0.4298  \\
\bottomrule
\end{tabular}%
}
\end{table}

\begin{table}[h]
\centering
\caption{\texttt{CovidQA} Dataset Results}
\label{tab:covidqa_results}
\resizebox{\columnwidth}{!}{%
\begin{tabular}{lccc}
\toprule
\textbf{Model/Optimizer} & \textbf{Micro F1} & \textbf{Macro F1} & \textbf{Weighted F1} \\
\midrule
Baseline GPT-4o                    & \textbf{0.9677}  & \textbf{0.9676}  & \textbf{0.9677}  \\
Baseline CoT DSPy                  & 0.8065  & 0.2482  & 0.8710  \\
Bootstrap Few Shot Random Search   & 0.9355  & 0.9354  & 0.9355  \\
Bootstrap Few Shot Optuna          & 0.9032  & 0.9028  & 0.9030  \\
MIPROv2                            & 0.9032  & 0.9028  & 0.9030  \\
KNN Few Shot                       & 0.9355  & 0.6343  & 0.9511  \\
COPRO                              & 0.8065  & 0.2482  & 0.8710  \\
RAGAS                              & 0.6452  & 0.6392  & 0.6377  \\
DeepEval                           & 0.6129  & 0.5303  & 0.5366  \\
\bottomrule
\end{tabular}%
}
\end{table}

\begin{table}[h]
\centering
\caption{\texttt{FinanceBench} Dataset Results}
\label{tab:financebench_results}
\resizebox{\columnwidth}{!}{%
\begin{tabular}{lccc}
\toprule
\textbf{Model/Optimizer} & \textbf{Micro F1} & \textbf{Macro F1} & \textbf{Weighted F1} \\
\midrule
Baseline GPT-4o                    & 0.6667  & 0.6667  & 0.6667  \\
Baseline CoT DSPy                  & 0.6667  & 0.4889  & 0.7333  \\
Bootstrap Few Shot Random Search   & 0.6667  & 0.6667  & 0.6667  \\
Bootstrap Few Shot Optuna          & \textbf{0.8333}  & \textbf{0.8286}  & \textbf{0.8286}  \\
MIPROv2                            & \textbf{0.8333}  & \textbf{0.8286}  & \textbf{0.8286}  \\
KNN Few Shot                       & 0.6667  & 0.6667  & 0.6667  \\
COPRO                              & 0.6667  & 0.6667  & 0.6667  \\
RAGAS                              & 0.5000  & 0.3333  & 0.3333  \\
DeepEval                           & 0.6667  & 0.6250  & 0.6250  \\
\bottomrule
\end{tabular}%
}
\end{table}

\end{document}